\documentclass[10pt,twocolumn,letterpaper]{article}

\usepackage{iccv}              %

\usepackage[accsupp]{axessibility}  %
\usepackage[dvipsnames]{xcolor}

\usepackage{multirow}
\usepackage{bbold} %
\usepackage{tabularx} %
\usepackage{colortbl}

\usepackage{amssymb}
\usepackage{algorithm}
\usepackage{algpseudocode}

\usepackage{soul}

\usepackage{pifont}
\newcommand{\xmark}{\ding{55}}%

\DeclareMathOperator*{\argmax}{arg\,max}

\definecolor{iccvblue}{rgb}{0.21,0.49,0.74}
\usepackage[pagebackref,breaklinks,colorlinks,allcolors=iccvblue]{hyperref}

\def\paperID{11556} %
\def\confName{ICCV}
\def\confYear{2025}

\title{Looking in the Mirror: A Faithful Counterfactual Explanation Method for Interpreting Deep Image Classification Models}

\author{Townim Faisal Chowdhury$^{1}$, 
Vu Minh Hieu Phan$^{1}$, 
Kewen Liao$^{2}$, 
Nanyu Dong$^{1}$, 
Minh-Son To$^{3}$, \\
Anton van den Hengel$^{1}$, 
Johan W. Verjans$^{1}$, 
\stepcounter{footnote}Zhibin Liao$^{1}$\thanks{Corresponding author.}\\
{\small $^{1}$Australian Institute for Machine Learning, University of Adelaide, Australia}, 
{\small $^{2}$Deakin University, Australia}, \\
{\small $^{3}$Flinders University, Australia}\\
}

\begin{document}

\maketitle

\begin{abstract}
Counterfactual explanations (CFE) for deep image classifiers aim to reveal how minimal input changes lead to different model decisions, providing critical insights for model interpretation and improvement. 
However, existing CFE methods often rely on additional image encoders and generative models to create plausible images, neglecting the classifier's own feature space and decision boundaries.
As such, they do not explain the intrinsic feature space and decision boundaries learned by the classifier.
To address this limitation, we propose Mirror-CFE, a novel method that generates faithful counterfactual explanations by operating directly in the classifier's feature space, treating decision boundaries as mirrors that ``reflect'' feature representations in the mirror. 
Mirror-CFE learns a mapping function from feature space to image space while preserving distance relationships, enabling smooth transitions between source images and their counterfactuals. 
Through extensive experiments on four image datasets, we demonstrate that Mirror-CFE achieves superior performance in validity while maintaining input resemblance compared to state-of-the-art explanation methods. 
Finally, mirror-CFE provides interpretable visualization of the classifier's decision process by generating step-wise transitions that reveal how features evolve as classification confidence changes. Code is available at: \textbf{\texttt{\url{https://github.com/AIML-MED/Mirror-CFE}}}.
\end{abstract}

\section{Introduction}

A transparent and human-understandable AI decision-making process is highly appreciated by stakeholders, policymakers, and the public to trust and use AI-supplied decisions, thus promoting AI adaptation and ensuring AI robustness and fairness. 
Explainable AI (XAI) methods generally attempt to demonstrate important input features \cite{zhou2016learning,chowdhury2024cape,ribeiro2016should,scott2017unified} to provide insight, yet deep learning image analysis models are extremely complex, making it difficult for researchers to formulate a way to unravel the decision process.
Attribution maps \cite{fong2017interpretable,chattopadhay2018grad}, concept-based explanations \cite{yuksekgonul2022post, koh2020concept}, and example-based explanations \cite{bien2011prototype,adhikari2019leafage} all aim to clarify the decision-making process of a model by associating the prediction outcome of the model with specific input features, image subregions, or referring to similar examples; yet they do not provide actionable insights \cite{rudin2019stop,adebayo2018sanity} and cannot explain model decisions if the illustrated patterns do not exist or are slightly altered in the first place.
On the other hand, imitating how humans (especially children \cite{beck2009relating, buchsbaum2012power}) learn through comparing likelihoods of events, Counterfactual Explanations (CFEs) \cite{wachter2017counterfactual, lucic2020does} have emerged as an effective solution, offering actionable insights by making minimal changes to the input that may lead to different model decisions. 
By modifying the original input to the model, counterfactual explanations can answer what-if questions, such as, "What if the patient has disease X, what would I see on their brain MRI?". CFEs not only enhance human comprehension \cite{miller2019explanation} but also allow users to "play" with AI recommendations by considering and weighing alternative scenarios.

On the generation task of visual CFE for image classification models, administrating minimal changes that can trigger decision flip is tricky, where the minimal changes may not produce perceptible or meaningful visual counterfactual explanations. 
The first group of CFE methods finds CFE instances either by exhaustive search in image feature space \cite{goyal2019counterfactual} or by optimizing custom-designed objective functions to fulfil the properties of CFE \cite{dhurandhar2018explanations,wachter2017counterfactual}. 
However, these methods occasionally fail to converge and lead to no finding of CFE. 
The second group of CFE methods \cite{akula2020cocox, wang2020scout} finds the regions or concepts by selecting the most similar examples to which the input should change towards to get a different outcome. 
The second group of methods highly relies on a diverse and large pool of discriminative images to identify possible and small changes in the input, impacting the diversity and realism of generated CFE images. 
Both groups of methods are prone to generate imperceptible visual CFE images. 
To tackle this problem, recently, a third group of methods \cite{lang2021styex, augustin2022diffusion, wang2021explaingan, khorram2022c3lt} use generative models to generate CFE images. 
These methods commonly rely on a trained generator model to produce realistic images from random latent encodings. 
This approach inevitably requires a paired encoder (or reverse engineering) for the generator and conditions the generation process on features or probabilities computed by the classifier. As such, the generated CFE does not explicitly reflect how the classifier decides, \ie, the CFE generation does not correlate to the \textit{actual} decision boundaries of the classifier.

In this paper, our goal is to generate faithful and plausible CFEs that align with the classifier's decision-making process. 
Additionally, we address an important aspect of CFE that remains unsolved in the literature; namely, current CFE methods only generate a single example that flips the classifier’s decision. Such discrete generation does not provide sufficient insight for users to understand "how can we make this change happen?". In a healthcare scenario, we believe it is more useful for clinicians to see a smooth transition from the input image to an altered counterpart that shows, "the patient would have disease X if these regions start to show pathological change on which brain regions".

\begin{figure}
    \centering
    \includegraphics[width=\columnwidth]{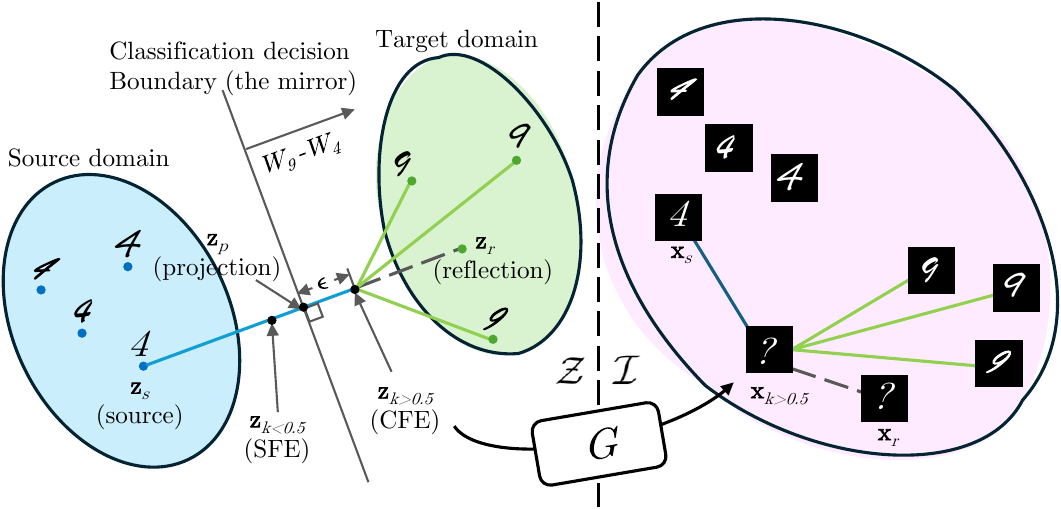}
    \caption{Illustration of the proposed Mirror-CFE method using an MNIST-like example that generates a CFE of 9 from a 4. In a nutshell, Mirror-CFE treats the decision boundary between the source and target classes (defined by $\mathbf{W}_9 - \mathbf{W}_4$) as a mirror in the classifier's feature space $\mathcal{Z}$. Our method finds the projection $\mathbf{z}_p$ of the source $\mathbf{z}_s$ on the mirror surface and the reflection $\mathbf{z}_r$ ``in'' the mirror.
    To maintain the sparsity property, our CFE is defined as a point $\mathbf{z}_{k}$ traveling on the line from $\mathbf{z}_s$ to $\mathbf{z}_r$, passing $\mathbf{z}_p$ by an arbitrary distance $\epsilon$ so the classifier would start to predict it as a 9.
    Similarly, Mirror-CFE can also generate a semi-factual explanation (SFE, opposite of CFE which does not change class prediction) without letting $\mathbf{z}_k$ pass $\mathbf{z}_p$.
    A unique design of Mirror-CFE is that our mapping function $G: \mathcal{Z} \rightarrow \mathcal{I}$ captures certain distance relationships (\eg, the ratio between the blue line and each of the green lines) in $\mathcal{Z}$ and maintain that in $\mathcal{I}$.
    }
    \label{fig:intro}
\end{figure}

Taking these objectives into consideration, we propose a novel method, Mirror-CFE, for faithful, plausible, and animated CFE generation.
As the name suggests, our method first finds a ``mirror'' in the classifier's feature space, which is computed by the classification boundary between the source class and the target class (the class we want to generate CFE).
Intuitively, our MirrorCFE rethinks counterfactual explanation (CFE) generation as a journey through the classifier's decision boundary—learning a mapping that generates images at every point along the path from the source image to its reflection. 
This enables us to animate the gradual transformation of the image as it transitions from the source class to the target class.
An example of Mirror-CFE is shown in Fig.~\ref{fig:intro}.
Finally, the optimization goal of Mirror-CFE is to find a mapping process that preserves certain distance relationships in the latent space to the image space, such as the scale difference between the blue and the green segments shown in Fig.~\ref{fig:intro}.
In summary, our main contributions are as follows:

\begin{itemize}
    \item Mirror-CFE is a novel approach that redesigns the CFE generation process, \ie, we compute the most plausible CFE point location in the classifier's latent space and focus on generating their counterparts in the image space.
    \item We design a triangulation loss function that leverages the distance relationship in the latent space to regularize the mapping function to maintain such a relationship in the image space. This enables us to generate the animated source image to CFE image transition.
    \item In Mirror-CFE, the classifier serves as the sole feature encoder and decision maker and is never fine-tuned, thereby upholding the explanation faithfulness to the classifier's decision-making boundary.
\end{itemize}

\section{Related Works}
\subsection{DNN Explanation} 
The most prominent way for explaining deep neural networks (DNN) is attribution-based methods, \ie, saliency based \cite{chattopadhay2018grad,jiang2021layercam,zhou2016learning,chowdhury2024cape} or perturbation based \cite{petsiuk1806rise, fong2017interpretable, zintgraf2017visualizing, chang2018explaining, ribeiro2016should} explanation methods. 
In general, attribution methods rely on the backpropagation algorithm or input perturbations to identify the image regions that are most important to a model’s decision. 
Saliency-based approaches, \eg, Class activation maps (CAM) \cite{zhou2016learning} and Grad-CAM \cite{chattopadhay2018grad} highlight spatial regions by examining global activation patterns from deep layers in DNN. 
However, CAM methods tend to highlight similar regions for similar semantic classes, limiting their actionable analytical capabilities. 
To address such an issue, a probabilistic normalization of CAM~\cite{chowdhury2024cape} among the classes was introduced. 
Nevertheless, saliency-based methods remain good sanity checks of models but provide limited interpretability and actionable insights~\cite{rudin2019stop,adebayo2018sanity}. 
Perturbation methods, \eg, adding noise \cite{chang2018explaining} or checking occlusion sensitivity \cite{ma2013visualizing}, associate a model's decision-making process with input changes.
Fong \etal \cite{fong2017interpretable} proposed meaningful perturbation to the parts of the input data, such as pixels in an image or words in the text to identify their importance in the model's decision. 
In addition, LIME \cite{ribeiro2016should} and SHAP \cite{scott2017unified} are popular model-agnostic perturbation methods that do not assume knowledge of the model architecture. 
However, perturbation methods sample the input space, forming a retrospective examination of what action caused the decision change, which makes it difficult to use them to conduct prospective explanations, \ie, if I want to make certain changes in the decision, what should the input be?

\subsection{Counterfactual Explanation} 
Counterfactual explanation (CFE) methods for machine learning models \cite{wachter2017counterfactual,goyal2019counterfactual,wang2020scout,dhurandhar2018explanations} investigate a model's possible output in relation to hypothetical circumstances. 
For example, a CFE transforms an input image of digit `3' to `8' (hypothetically) in order to make a digit classifier consider it `8'.
CFE has been traditionally studied on structured data to gain actionable insights from data \cite{wachter2017counterfactual, mothilal2020explaining, white2020measurable, pawelczyk2020learning}.
Recently, researches have started applying CFEs on visual explanations \cite{goyal2019counterfactual, lang2021styex, dhurandhar2018explanations}.

Optimization-based methods \cite{dhurandhar2018explanations,khorram2022c3lt,joshi2021decompose,kenny2021generating} generate CFEs by designing objective functions within an optimization framework. Recent works \cite{khorram2022c3lt,joshi2021decompose,kenny2021generating} use pretrained DNNs to guide generation. Joshi \etal \cite{joshi2021decompose} extend causal analysis by optimizing in the latent space of a pretrained VAE. Kenny \etal \cite{kenny2021generating} propose PIECE, using a GAN to modify salient latent features and produce contrastive explanations. Khorram \etal \cite{khorram2022c3lt} transform the latent space of a conditioned decoder, but may introduce bias from an external encoder not aligned with the target classifier. As these methods depend on pretrained decoders, the resulting CFEs can diverge from the input image \cite{holtgen2021deduce}, making them harder to interpret .

In contrast, another CFE approach \cite{goyal2019counterfactual,akula2020cocox,wang2020scout,vandenhende2022making} focuses on replacing specific image regions from a target image. 
Goyal \etal \cite{goyal2019counterfactual} first proposed a greedy, sequential, exhaustive search within the classifier’s feature space to replace key regions of the input image. Vandenhende \etal \cite{vandenhende2022making} refined this by focusing on identifying and replacing only the most semantically meaningful region to generate counterfactual instances. 
However, CFE images from this type of method often result in less human-interpretable instances and they frequently fail to converge \cite{dhurandhar2018explanations,goyal2017counterfactual}.

On the visual plausibility of CFE, research \cite{lang2021styex, augustin2022diffusion, wang2021explaingan} has used generative models \cite{goodfellow2014generative, ho2020denoising} to address plausibility issues in CFE generation. 
\cite{lang2021styex} utilized StyleGAN2 \cite{karras2020analyzing} to discover class attributes through the style space. 
Wang \etal \cite{wang2021explaingan} employed a generative approach to create realistic counterfactual images by filling a masked region of the input. 
Recent methods \cite{augustin2022diffusion, jeanneret2022diffusion} employed diffusion models to generate more realistic CFEs. 
Although these methods focus on realism, they are resource-intensive to train and often lead to unfaithful explanations \cite{dhurandhar2018explanations}, \ie, they don't align with the exact model decision logic. 

At last, the field is still investigating which requirements are necessary for a good counterfactual explanation. 
GdVAE~\cite{haselhoff2024gaussian} is a recent self-explainable model that integrates a conditional variational autoencoder with a Gaussian discriminant analysis classifier. Guidotti \etal \cite{guidotti2024counterfactual} outlined essential requirements for counterfactual explanations, while \cite{altmeyer2024faithful} emphasized prioritizing faithfulness to the model’s decision-making process over plausibility to ensure meaningful and accurate CFE to the investigated model.

\section{Definitions} 

Let $\mathbf{x} \in \mathcal{X}$ be a single image and $y \in \mathcal{C}$ be the corresponding label, where $\mathcal{X}$ and $\mathcal{C}$ denote the image and label sets respectively. 
We denote the image encoder function of a classifier as $\mathbf{z} = F(\mathbf{x}) \in \mathbb{R}^N$ where $F$ denotes the feature encoder, and $\mathbf{z} \in \mathcal{Z}$ denotes $\mathbf{x}$'s counterpart in the latent feature space $\mathcal{Z}$. 
A classification layer sits on top of $F$, producing a probability distribution $\mathbf{p} = \sigma(\mathbf{z}) = \text{softmax}(\mathbf{W}^\top\mathbf{z} + \mathbf{b}), \mathbf{p} \in \mathbb{R}^{|\mathcal{C}|}$, where $\mathbf{W} \in \mathbb{R}^{N \times |\mathcal{C}|}$ and $\mathbf{b} \in \mathbb{R}^{|\mathcal{C}|}$ are the weights and bias. 
Table~\ref{tab:symbol_lookup} in the supplementary material collects the full list of symbols defined in this paper.

Given a source class ($s$) image $\mathbf{x}_s$, a generated CFE image $\mathbf{x}_{cf}$ that changes the classifier decision to a target class ($t$) should meet desired CFE properties~\cite{moraffah2020causal, guidotti2024counterfactual, verma2024counterfactual}, which are summarized and explained below.

\begin{enumerate}
    \item \textit{Validity.} The prediction of the CFE matches the target class, \ie, $\argmax{\mathbf{p}_{cf}} = t$. 
    \item \textit{Realism (Plausibility).} $\mathbf{x}_{cf}$ should look like a real image from $\mathcal{X} \subset \mathcal{I}$ where $\mathcal{I}$ represents the image (pixel) space.
    \item \textit{Proximity (Similarity).} $\mathbf{x}_{cf}$ should be close to $\mathbf{x}_s$ for defined distance functions.
    \item \textit{Sparsity (Minimality).} Multiple interpretations of sparsity~\cite{guidotti2024counterfactual, ma2013visualizing, zhang2018unreasonable, bengio2013representation} exist in the literature.
    In brief, sparsity can be thought of as having minimal change in the features (attributes) between $\mathbf{x}_{cf}$ and $ \mathbf{x}_s$.
    Here we follow the definition in~\cite{zhang2018unreasonable} which compares $\mathbf{x}_{cf}$ and $ \mathbf{x}_s$ in an abstracted feature space. 
    \item \textit{Diversity.} Given multiple CFEs exist for explaining one source image, the CFEs should satisfy the other CFE properties while remaining different from each other.
   \item \textit{Faithfulness.} By definition in~\cite{altmeyer2024faithful}, faithfulness is the degree to which a CFE is consistent with what the classifier has learned about the data $\mathcal{X}$. Our interpretation is that a classifier coarsely learns two things, a feature embedding and a set of classification boundaries, hence a faithful CFE method should construct CFEs that meaningfully correspond to both.
\end{enumerate}

\section{Methodology}

Our Mirror-CFE answers to all six CFE properties.
In particular, Mirror-CFE supports a faithful CFE generation by making explicit explanations adhere to the classifier's decision process.

\subsection{Step-based CFE generation in latent space}

To achieve the faithfulness property, we first define what should be considered a CFE in the classifier's latent space $\mathcal{Z}$. 
To illustrate that, a two-class classification example is given in Fig.~\ref{fig:intro}.
We first compute the pairwise decision boundary between the source class $s$ and target class $t$, which is a hyperplane parameterized by the difference of their class weights $\mathbf{W}_m=\mathbf{W}_t - \mathbf{W}_s$, for $\mathbf{W}_s, \mathbf{W}_t \in \mathbb{R}^N$ denotes the vertical ``slices'' of $\mathbf{W}$ corresponding to $s$ and $t$, and bias $\mathbf{b}_m = \mathbf{b}_t - \mathbf{b}_s$, for $\mathbf{b}_s, \mathbf{b}_t$ denoting the elements of $\mathbf{b}$.
For simplicity, we call the decision boundary a ``mirror''.
Then, we define a position function:
\begin{equation}
    P(\mathbf{z}_s, \mathbf{W}_m, \mathbf{b}_m, k) = \mathbf{z}_s - 2k (\mathbf{W}^\top_m \mathbf{z}_s + \mathbf{b}_m) \hat{\mathbf{W}}_m,
    \label{eq:position}
\end{equation}
where $\hat{\mathbf{W}}_m = \frac{\mathbf{W}_m}{\|\mathbf{W}_m\|}$ represents the unit vector, $k \in [0, 1]$ is a step factor.
With $P$, we can ``travel'' $\mathbf{z}_s$ on the direction of $\hat{\mathbf{W}}_m$ by $k = 0.5$ so that $\mathbf{z}_s$ reaches the surface of the mirror and geometrically become its projection $\mathbf{z}_p$; or $k = 1$ so that it become its reflection $\mathbf{z}_r$ ``in'' the mirror. 
The property of $\mathbf{z}_p$ and $\mathbf{z}_r$ are that they are the closest points in $\mathcal{Z}$ which computes the equal probability (\ie, $\mathbf{q}_{(p,s)} = \mathbf{q}_{(p,t)}$, the subscripts denote the point name and the class) and the flipped class confidence (\ie, $\mathbf{q}_{(r,s)} = \mathbf{q}_{(s,t)}$ and vice versa), computed by a pairwise two-class classification $\mathbf{q} = \text{sigmoid}(\mathbf{W}_m^\top \mathbf{z} + \mathbf{b}_m)$.

\textbf{CFE definition in $\mathcal{Z}$}. 
Benefiting from Eq.~(\ref{eq:position}), we define a point $\mathbf{z}_{cf}$ is a CFE for $\mathbf{z}_s$ if it is on the line segment between $\mathbf{z}_r$ to $\mathbf{z}_p$ and $\mathbf{q}_{(cf, t)} > \mathbf{q}_{(cf, s)}$.
This equivalents to a step factor $k = 0.5 + \epsilon$ where $0<\epsilon <=0.5$ in Eq.~(\ref{eq:position}).
Our CFE definition answers the following CFE properties: (1) \textit{Validity}. $\mathbf{z}_{cf}$ achieves the change of classification to the target class.
(2) \textit{Sparsity}. $\mathbf{z}_{cf}$ is the closest point to $\mathbf{z}_s$ which achieves a specific probability distribution $\mathbf{q}_{cf}$.
(3) \textit{Diversity}. $\mathbf{q}_c$ can be changed to meet a required confidence level by changing $k$, \ie, we can generate a CFE that gives 0.6 probability of the target class as well as a CFE for 0.7 probability.
Finally, we do not consider CFE beyond $k=1$ as moving further away from $\mathbf{z}_s$ reduces the CFE's significance and challenges the sparsity property.

\textbf{Semi-factual explanations (SFE).}
Intuitively, Eq.~(\ref{eq:position}) can also generate SFE, noted as $\mathbf{z}_{sf}$, which processes a subtle change to $\mathbf{z}_s$ and characterized by $k = 0.5 - \epsilon$.
Our SFE definition also supports the validity (for SFE, it means keeping the same classification outcome), sparsity, and diversity properties.

\textbf{Step-based factual explanations (KFE).}
Given CFE and SFE are distinguished by $k$ above/below 0.5, we name the union of CFE and SFE step-based factual explanations (KFE, K refers to the step factor $k$).

Finally, Mirror-CFE is applicable to multi-class classification models but requires estimating the reflection point's location, please refer to Sec.~\ref{sec:multiclass_mirror_cfe} in the supplementary material for the calculation.
 
\subsection{Mapping from latent space to image space}

After we defined the KFE points in $\mathbf{z}$, our goal is to find a mapping function $G$ to project these points to the pixel space $\mathcal{I}$, \ie, $\mathbf{x} = G(\mathbf{z}): \mathcal{Z} \rightarrow \mathcal{I}$.
The goal of $G$ is to maintain the validity, proximity, and realism during the mapping process.
Therefore, we design the following loss functions to regularize $G$ to fulfill these objectives.
We note that the loss functions, except our proposed triangulation loss, are commonly used in GAN literature.

\textbf{Classification loss.} To enforce the CFE validity, a generated KFE $\mathbf{x}_k$ should achieve the same classification outcome as their latent space counterpart $\mathbf{z}_k$ had.
This validity can be generalized to the real samples if they have been featurized first:
\begin{equation}
\mathcal{L}_\text{cls} = \mathbb{E}_{\mathbf{z}_k \in p_\text{kfe}, \mathbf{x} \in \mathcal{X}}[\text{KLD}(\hat{\mathbf{p}}, \mathbf{p})],  
\label{eq:cls}
\end{equation}
where $p_\text{kfe}$ denotes the KFE distribution in $\mathcal{Z}$, KLD denotes the Kullback-Leibler divergence loss, $\hat{\mathbf{p}} = \sigma\big( F(G(\mathbf{z}))\big)$ for $\mathbf{z} \in \{F(\mathbf{x}), \mathbf{z}_k\}$, $\mathbf{p} = \sigma(\mathbf{z})$ for $\mathbf{z} \in \{F(\mathbf{x}), \mathbf{z}_k\}$.  
We use KLD to promote a finer source and target class confidence flip which contrasts with the common usage of cross-entropy loss in the literature \cite{khorram2022c3lt,wang2021explaingan} that only ensures a flip in the classification outcome.

\textbf{Adversarial loss.} 
A KFE needs to look real hence we follow the literature to use an Adversarial loss \cite{goodfellow2020generative} to promote the realism property:
\begin{equation}
    \mathcal{L}_\text{adv} = \mathbb{E}_{x \sim\mathcal{X}}[\text{log}(D(\mathbf{x}))] + \mathbb{E}_{\mathbf{z}_k \sim p_\text{kfe} } [\text{log}(1 - D(G(\mathbf{z}_k)))],   
\end{equation}
where the generator $G$ tries to minimize $\mathcal{L}_\text{adv}$ while a discriminator $D$ tries to maximize it.

\textbf{Reconstruction loss.} The reconstruction loss (or cycle consistency loss \cite{zhou2016learning, godard2017unsupervised}) is an indirect support to the validity by making sure that after being featurized and regenerated, the regenerated counterparts will look like the real image themselves and hence will help the optimization of Eq.~(\ref{eq:cls}) for the real images:
\begin{equation}
\mathcal{L}_\text{rec} = \mathbb{E}_{\mathbf{x} \sim \mathcal{X}}[|\mathbf{x} - G(F(\mathbf{x}))|].    
\end{equation}

\textbf{Feature reconstruction loss.}
The above reconstruction loss regularizes for the real images. 
To promote the cycle consistency for the KFEs, we similarly design a feature reconstruction loss to aid validity using the L2 distance:
\begin{equation}
    \mathcal{L}_\text{fea} = \mathbb{E}_{\mathbf{z}_k \sim p_\text{cfe}}[\|\mathbf{z}_k - F(G(\mathbf{z}_k))\|].
\label{eq:fea}
\end{equation}

\textbf{Triangulation loss.}
In literature, a proximity loss is often used to make sure a CFE image would look close to the source image, therefore $\mathcal{L}_\text{prox} = |\mathbf{x}_s - \mathbf{x}_{cf}|$.
In practice, we found using $\mathcal{L}_\text{prox}$ makes KFEs $\mathbf{x}_{k>0.5}$ visually resemble $\mathbf{x}_s$ but still being classified as the target class, \ie, $G$ starts to generate adversarial samples by adding visually imperceptible noises to change the classification outcome.
This behavior means $G$ learns to find a trivial solution that satisfies the CFE properties but it reduces plausibility.
In order to address this plausibility issue, we try to maintain the pixel space relationship (L1) between a CFE image and the real target images according to their semantic relationship (L2) provided by their latent space counterparts:
\begin{equation}
    \frac{|\mathbf{x}_k - \mathbf{x}_t|}{|\mathbf{x}_s - \mathbf{x}_k|} \approx
    \frac{\|\mathbf{z}_k - \mathbf{z}_t\|}{\|\mathbf{z}_s - \mathbf{z}_k\|} = \beta,
\end{equation}
for any $\mathbf{x}_t \in \mathcal{X}$ and for $k = 0.5 + \epsilon$. Then we introduce a relaxation which derives the lower and upper bound for the distance between $|\mathbf{x}_s - \mathbf{x}_k|$:
\begin{equation}
    (1-\alpha)\frac{|\mathbf{x}_k - \mathbf{x}_t|}{\beta} \leq |\mathbf{x}_s - \mathbf{x}_k| \leq (1+\alpha)\frac{|\mathbf{x}_k - \mathbf{x}_t|}{\beta},
\end{equation}
where $\alpha \in [0, 1]$ and the loss term is implemented as:
\begin{align}
    \mathcal{L}_\text{cfe} =  & \max(\frac{(1-\alpha)}{\beta}|\mathbf{x}_k - \mathbf{x}_t| - |\mathbf{x}_s - \mathbf{x}_k|, 0)\nonumber \\
    + & \max(|\mathbf{x}_s - \mathbf{x}_k| - \frac{(1+\alpha)}{\beta}|\mathbf{x}_k - \mathbf{x}_t|, 0),
\label{eq:cfe}
\end{align}
for $k = 0.5 + \epsilon$.
As Eq.~(\ref{eq:cfe}) is designed to regularize only the CFE generation, it leaves the generations on the segment of $k < 0.5$ (SFE segment) unattended.
Following the same design of $\mathcal{L}_\text{cfe}$, we spell out the symmetric loss which maintains the distance relationship between any SFE $\mathbf{x}_{sf}$ to a random source class images $\mathbf{x}_{ss}$ (including $\mathbf{x}_{s}$):
\begin{equation}
    \frac{|\mathbf{x}_{k} - \mathbf{x}_{ss}|}{|\mathbf{x}_s - \mathbf{x}_{k}|} \approx
    \frac{\|\mathbf{z}_{k} - \mathbf{z}_{ss}\|}{\|\mathbf{z}_s - \mathbf{z}_{k}\|} = \gamma,
\end{equation}
and
\begin{align}
    \mathcal{L}_\text{sfe} =  & \max(\frac{(1-\alpha)}{\gamma}|\mathbf{x}_{k} - \mathbf{x}_{ss}| - |\mathbf{x}_{s} - \mathbf{x}_{k}|, 0)\nonumber \\
    + & \max(|\mathbf{x}_{s} - \mathbf{x}_{k}| - \frac{(1+\alpha)}{\gamma}|\mathbf{x}_{k} - \mathbf{x}_{ss}|, 0),
\label{eq:sfe}
\end{align}
for $k = 0.5 - \epsilon$.
Finally, our triangulation loss merges the above into one:
\begin{equation}
    \mathcal{L}_\text{tri} = \mathbb{1}(k < 0.5) \mathcal{L}_\text{sfe} + \mathbb{1}(k >= 0.5) \mathcal{L}_\text{cfe},
\end{equation}
for any $k \in [0, 1]$, where $\mathbb{1}$ denotes the indicator function.
The naming illustrates the similarity of this process to the process of determining the location of an airplane (\ie, an SFE or CFE image's image space coordinates) by measuring its distances to known satellites (the source and target images' image space coordinates).

\subsection{Model design}

\begin{figure*}
    \centering
    \includegraphics[width=0.75\textwidth]{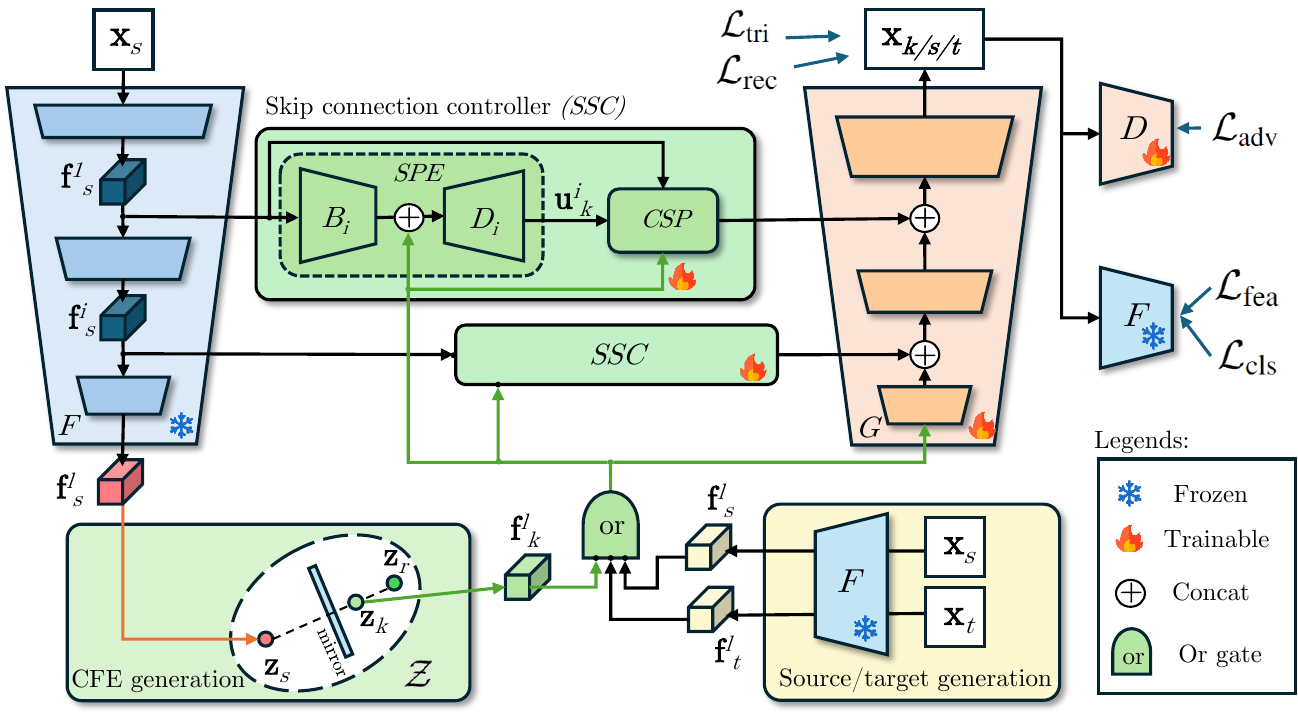}
    \caption{An overview of the Mirror-CFE architecture and workflow. During the training, we allow the input of the skip connection controller (SSC) module to take either $\mathbf{f}_k^l$ to generate an SFE or CFE or $\mathbf{f}_{s/t}^l$ to generate a source/target image. The `or' gate indicates the architecture can also handle source and target reconstruction. The main contributions of the Mirror-CFE are (1) the CFE generation process in $\mathcal{Z}$, (2) the SSC module which allows control of the information flow from the encoder $F$ to decoder $G$, (3) the triangulation loss which regularizes the CFE image $\mathbf{x}_k$ and (4) the ability to move $\mathbf{z}_k$ along $\mathbf{z}_s$ to $\mathbf{z}_r$ to generate an animated CFE transition. }
    \label{fig:method}
\end{figure*}

Mirror-CFE employs an encoder-decoder architecture to facilitate image encoding and reconstruction.
Instead of reverse engineering a separate encoder $E$ like the prior methods \cite{khorram2022c3lt, lang2021styex}, we utilize $F$ of the classifier as $E$ and do not train $F$ to maintain faithfulness.
The encoder-decoder structure works well for the low-resolution MNIST and FMNIST datasets, but the generated image becomes blurry or artificial for high-resolution datasets.
This suggests the high-frequency information was not captured by the classifier's encoding.
However, to remain a faithful CFE generation method, we must use $F$ as the encoder.

To solve this problem, we first employ the U-Net~\cite{ronneberger2015u} like skip connections to allow the high-frequency information to pass from $F$ to $G$.
The model architecture is shown in Fig.~\ref{fig:method}.
Formally, we define a set of features $\{\mathbf{f}^i\}_{i=1}^{l} \in \mathbb{R}^{C_{i}\times H_{i} \times W_{i}}$ from selected layers of $F$, which are passed to $G$ on $i$-th skip connection, here $l$ represents the last layer.
Each channel in $\mathbf{f}^i$ may represent a ``style'' information, \eg, pattern, texture, color, \etc. 
Because the skip connections would allow those styles to bypass $\mathbf{z}$'s control of the $G$, we propose a skip connection controller (SCC) module that edits the features before passing to $G$.

\textbf{Skip connection controller (SSC).} The SSC module has two components: a spatial pattern editor module, and a CAM-guided KFE localization module.

\textit{(1) Spatial pattern editor (SPE) module.}
The SPE module is designed to tap the skip connections passing from $F$ to $G$, realizing the KFE latent encoding $\mathbf{z}_k$'s control privilege to high-frequency style information.
Let $\mathbf{f}_{k}^{l}$ denote the last layer features from $\mathbf{z}_k$. $\mathbf{f}_{k}^{l}$ can be estimated from $\mathbf{z}_k$ and the calculation is given in Sec.~\ref{sec:kfe_enc_from_latent} in the supplementary material.
During the training, $\mathbf{f}_{k}^{l}$ can also be $\mathbf{f}_{s}^{l}$ or $\mathbf{f}_{t}^{l}$ to facilitate the source/target reconstruction jointly.
Lastly, we note $\mathbf{f}_{s}^{i}$ as the source image's feature at a layer $i \neq l$.
To integrate these features in the SPE module, the source skip connection feature $\mathbf{f}_{s}^{i}$ is first encoded via a bottleneck model $B_i$ to match the shape of $\mathbf{f}_{k}^{l}$ and then concatenate with $\mathbf{f}_{k}^{l}$ before it is projected back to original size by a decoder model $D_i$:  

\begin{equation}
\mathbf{u}_{k}^{i} = D_i\big(\text{concat}(B_i(\mathbf{f}_{s}^{i}),  \mathbf{f}_{k}^{l})\big),
\end{equation}
where $\mathbf{u}_{k}^{i} \in \mathbb{R}^{C_{i}\times H_{i} \times W_{i}}$ denotes the SPE module's output at $i$-th layer.

\textit{(2) CAM-guided spatial prior (CSP) module.} 
We leverage the Class Activation Map (CAM) information to obtain a spatial region, namely a \textit{spatial prior mask}, to indicate where changes should take place. 
The motivation of CSP is to promote the proximity between a KFE image to the source image by modifying most class discriminative regions as illustrated by CAM.
To briefly describe the calculation, unnormalized CAMs can be computed by directly applying the classification weights $\mathbf{W} \in \mathbb{R}^{N\times|\mathcal{C}|}$ to $\mathbf{f}^{l}_{k}$ without first engaging the Global Average Pooling layer, \ie, $\mathbf{U}_k = \mathbf{W}^\top \mathbf{f}^{l}_{k} \in \mathbb{R}^{|\mathcal{C}|\times H_l \times W_l}$ and the normalized CAMs are computed as $\mathbf{N}^c_k = \frac{\max(\mathbf{U}^c_k, 0)}{\max(\mathbf{U}^c_k)}$ for an arbitrary class $c$.
Then we utilize the source and target class CAMs and take the union on their binarized copies to compose the spatial prior mask:
\begin{equation}
    \mathbf{M}^{i}_{k} = \mathbb{1}(\mathbf{N}^s_k > \rho ) \cup \mathbb{1}(\mathbf{N}^t_k > \rho), 
\end{equation}
where $\rho = \min(\max(1 - k, \rho^l), \rho^u)$ is a binarization threshold to determine the size of KFE spatial prior based on $k$ and bounded by user-supplied value range $[\rho^l, \rho^u]$.
Finally, we apply the spatial prior mask to control the spatial mixture of $\mathbf{f}^{i}_{s}$ and $\mathbf{u}^{i}_{k}$, \ie, 
\begin{equation}
\label{eq:mixture}
    \mathbf{f}'_{i} = (1-\mathbf{M}^{i}_{k})\times \mathbf{f}^{i}_{s} + \mathbf{M}^{i}_{k} \times \mathbf{u}^{i}_{k},
\end{equation}
where $\mathbf{f}'_{i}$ is then fed to $G$ in the corresponding layer.
In brief, as $k$ goes from 0 to 1, the size of the spatial prior mask increases to allow more spatial changes to be made.

\section{Experiments} 
We compare to five recently proposed CFE methods: REVISE \cite{joshi2021decompose}, CEM \cite{dhurandhar2018explanations}, ExpGAN \cite{wang2021explaingan}, PGD \cite{madry2019pgd}, and C3LT \cite{khorram2022c3lt} and employ the popular ResNets \cite{he2016deep} as the testing architecture. 
All experiments were conducted on a 48G NVidia RTX A6000 GPU using PyTorch \cite{NEURIPS2019_9015}.

\textbf{Datasets.}
\textbf{(1) MNIST}~\cite{lecun1998gradient} is a well-known dataset for hand-written digit classification containing digits 0 to 9. 
The dataset has 60,000 images for training and 10,000 for testing, the images are grayscale and in $28\times28$ pixel size.
\textbf{(2) Fashion-MNIST (F-MNIST)}~\cite{xiao2017fashion} is also a $28\times28$ grayscale image dataset consisting of 70,000 cloth and accessory images equally distributed in 10 categories. 
F-MNIST also has 60,000 images for training and 10,000 for testing.
\textbf{(3) Blood-MNIST (B-MNIST)}~\cite{yang2023medmnist} comprises 17,092 microscopic RGB cytology blood cell images in 8 classes, split into 11,959 images for training, 1,712 for validation, and 3,421 for testing. We utilize the $\mathbf{128\times128}$ image size for our experiment to compare the CFE methods at a larger spatial size.
\textbf{(4) CelebA-HQ}~\cite{karras2017progressive} comprises 30,000 high-quality celebrity faces, from which we select 10,000 training and 6,000 testing images based on the "slightly mouth open" attribute. To compare with C3LT, we use the $\mathbf{224\times224}$ image size for our experiment.

\textbf{Implementation details.}
We use ResNet-18 for MNIST and F-MNIST (modified to one input channel) and ResNet-50 classifier for B-MNIST and CelebA-HQ. 
For all four datasets, we use Adam \cite{kingma2014adam} optimizer with $2$e-4 learning rate. 
With these parameters, the ResNets achieve 99\%, 94\%, 99\%, and 84\% mean accuracy for the four datasets in the same order. 
We uniformly choose $\alpha = 0.2$ and use equal weights for all introduced loss terms except for $\mathcal{L}_\text{tri}$, a weight of 2 for F-MNIST and B-MNIST.
We evaluate the same class pairs defined in~\cite{wang2021explaingan} for the MNIST and F-MNIST datasets. For MNIST, the class pairs are 3 \vs 8, 4 \vs 9, and 5 \vs 6.
For F-MNIST, the class pairs are coat \vs shirt, T-shirt \vs pullover, and sneakers \vs boots. 
Finally, we select the erythroblast (an immature red blood cell) and lymphocyte (a type of white blood cell) for B-MNIST and the slightly open \vs closed mouths for CelebA-HQ.

\begin{figure}[!tbp]
    \centering
    \includegraphics[width=\columnwidth]{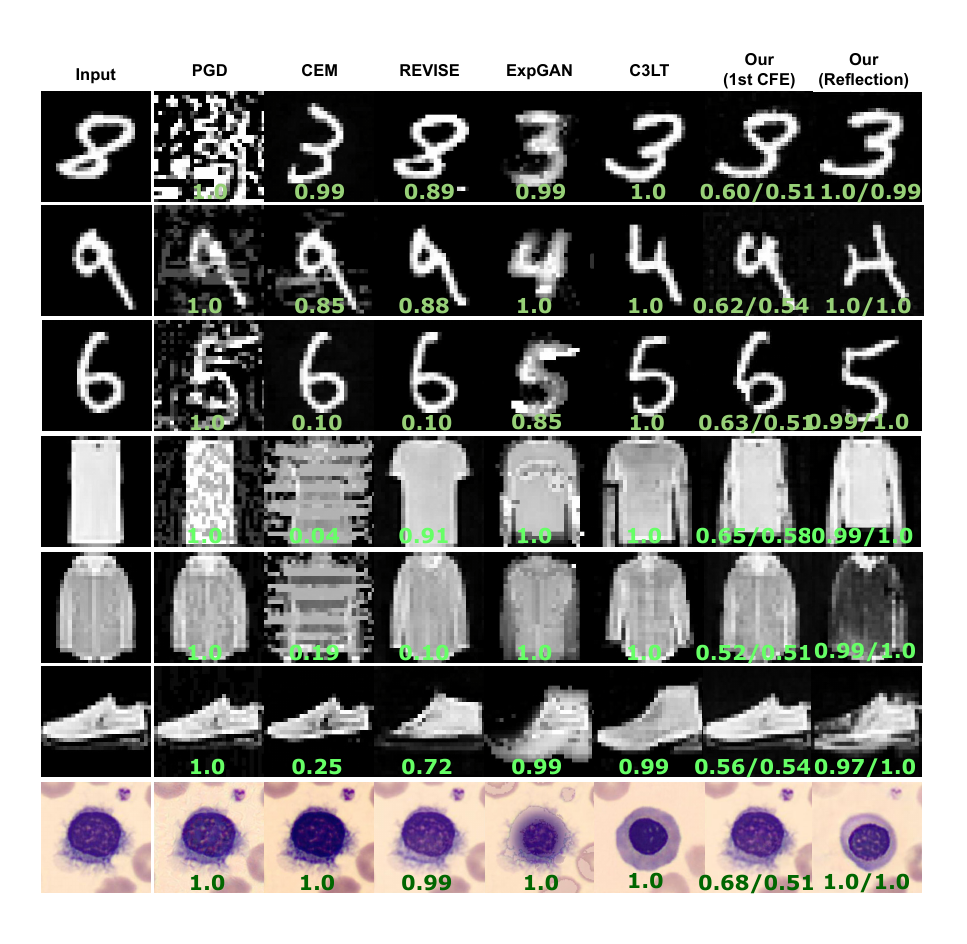}
    \caption{Qualitative CFE comparison for comparing methods. 1-3rd row: MNIST (8 $\rightarrow$ 3, 9 $\rightarrow$ 4, 6 $\rightarrow$ 5), 4-6th row: F-MNIST (T-shirt $\rightarrow$ pullover, coat $\rightarrow$ shirt, sneaker $\rightarrow$ anker boot), 7th row: B-MNIST (lymphocyte$ \rightarrow$ erythroblast).
    The image sizes are $28\times28$ for MNIST and F-MNIST and $128\times128$ for B-MNIST.
    We show the predicted target class confidence at the bottom of each image.
    For Mirror-CFE, we show the intended target class confidence after ``/''.}
    \label{fig:qualitative}
\end{figure}

\begin{table*}[!tbp]
\centering
\resizebox{1.0\textwidth}{!}{
\begin{tabular}{c|c|c|c|c|c|c|c|c|c|c|c|c|c|c|c|c|c|c}
\toprule
Dataset $\rightarrow$ & \multicolumn{6}{c|}{MNIST} & \multicolumn{6}{c|}{F-MNIST} & \multicolumn{6}{c}{B-MNIST}\\
\hline
Method & L1$\downarrow$ & LPIPS$\downarrow$ & FID $\downarrow$& D.Val$\uparrow$ & Val.$\uparrow$ & \% Fail& 
L1$\downarrow$  & LPIPS$\downarrow$  & FID$\downarrow$ & D.Val$\uparrow$ & Val.$\uparrow$ & \% Fail & 
L1$\downarrow$  & LPIPS$\downarrow$ & FID$\downarrow$ & D.Val$\uparrow$ & Val.$\uparrow$ &\% Fail\\
\midrule
 PGD \cite{madry2019pgd} & 
 0.42 & 0.31 & 15.62 & \underline{0.74} & \bf 1.0 & 0.0 &
 0.34 & 0.28 & 10.12 & 0.88 & \bf 1.0 & 0.0 &
  0.11 & 31.68 & 147.59 & \bf 0.99 & \bf 0.99 & 0.0\\
 REVISE \cite{joshi2021decompose} & \bf 0.08 & \underline{0.15} & \bf 1.96 &	0.15 & 0.02 & 9.77 & 
 \underline{0.14} & \underline{0.16} & 5.65 & 0.08 & 0.19 & 0.83 & 
 \bf 0.05 & \underline{4.58} & 98.39 & 0.37 & 0.11 & 21.12\\
 CEM \cite{dhurandhar2018explanations} & 0.32 & \bf 0.06 & \underline{2.71} & 0.16 &	0.03 & 16.72 & 
 0.28 & 0.24 & 13.77 & 0.04 & 0.10 & 46.40 & 
 0.12 & \bf 2.25 & \bf 31.30 & 0.44 & 0.14 & 0.0\\
 ExpGAN \cite{wang2021explaingan} & 0.25 & 0.31 & 26.22 & 0.61 & \bf 1.0 & 0.0 & 
 0.33 & 0.31 & 14.72 & 0.82 & \underline{0.99} & 0.0 & 
 \underline{0.10} & 31.14 & 101.17 & 0.50 & \underline{0.60} & 0.0\\
 C3LT \cite{khorram2022c3lt}& 0.17 & 0.25 & 8.95 & 0.79 & \bf 1.0 & 0.0 & 
 0.32 & 0.30 & 11.55 & 0.87 & \bf 1.0 & 0.0 & 
 0.16 & 30.14 & 96.03 & \bf 0.99 & \bf 0.99 & 0.0\\
 \hline
  Our (1st CFE) & \underline{0.16} & 0.17 & 3.25 & \bf 0.99 & \bf 1.0 & 0.0 &
 \bf 0.12 &	\bf 0.10 & \bf 2.80 & \bf 0.99 & \underline{0.99} & 0.0 &
 \bf{0.05} & 11.81 & \underline{86.02} & \bf 0.99 & \bf 0.99 & 0.0 \\
 Our ($k=1$) & 0.26 & 0.33 & 3.20 & \bf 0.99 & \bf 1.0 & 0.0 &
 0.25  & 0.21 & \underline{4.40} & \underline{0.96} & \underline{0.99} & 0.0 &
 0.14 & 28.69 & 132.93 & \underline{0.98} & \bf{0.99} & 0.0 \\
 \bottomrule
\end{tabular}
}
\caption{Quantitative comparison on the benchmarked datasets. \% Fail indicates the percentage of test images a method cannot find CFE, which occasionally happens to optimization-based CFE methods. 1st place in each column is marked in bold font, 2nd place by underline.}
\label{tab:quantitative}
\end{table*}

\subsection{Qualitative Evaluations}

In Fig.~\ref{fig:qualitative}, we show the qualitative comparison of 7 selected examples.
In general, we found that CFE generation is prone to generating adversarial example generation.
Indeed the two tasks have close objectives. 
CFE aims to find small but most meaningful changes to explain a decision flip, while adversarial generation aims to find small and imperceptible changes to trick an alternative decision.
In our opinion, we found PGD exhibits noise overlaying behavior, which fails at realism, and CEM is prone to adversarial generation.
REVISE generates better CFE examples for F-MNIST but again generates more adversarial-like examples for MNIST and B-MNIST.
On the contrary, ExpGAN gives target-like CFEs but they do not resemble inputs.
C3LT generates the most sensible CFEs across the five compared methods, balancing the visual resemblance to input and target-alike visual characteristics.
Finally, we show our Mirror-CFE in two settings: (1) \textit{1st CFE}, a CFE example intermediate after the classification decision is flipped (entering the mirror), and (2) the reflection point's CFE ($k=1$).
The 1st CFE shows a good overall resemblance while making small but meaningful ``line rubbing'' edits for all MNIST examples. 
For F-MNIST, 1st CFE gives adversarial examples except for the T-shirt $\rightarrow$ pullover, which adds the sleeves while maintaining the T-shirt's overall structure.
For B-MNIST, the 1st CFE gives an adversarial example.
As the $k$ increases to 1, Mirror-CFE breaks away from the adversarial effects, allowing more spatial patterns to be changed to target-like while increasing the target confidence.
To conclude, we found Mirror-CFE generation has the ability to control the number of changes to be adversarial or CFE. Also it gives users a sense of what's the most critical change between the two classes (happens earlier in the transition) and what's good to have (happens late in the transition).
Finally, we show the intended (\ie, $\mathbf{z}_k$ computed) target class confidence on the right side of predicted target class confidence, illustrating the degree of the faithfulness of the generation (also see Table.~\ref{tab:faithfulness}).

\begin{figure}[!bp]
    \centering
    \includegraphics[width=\columnwidth]{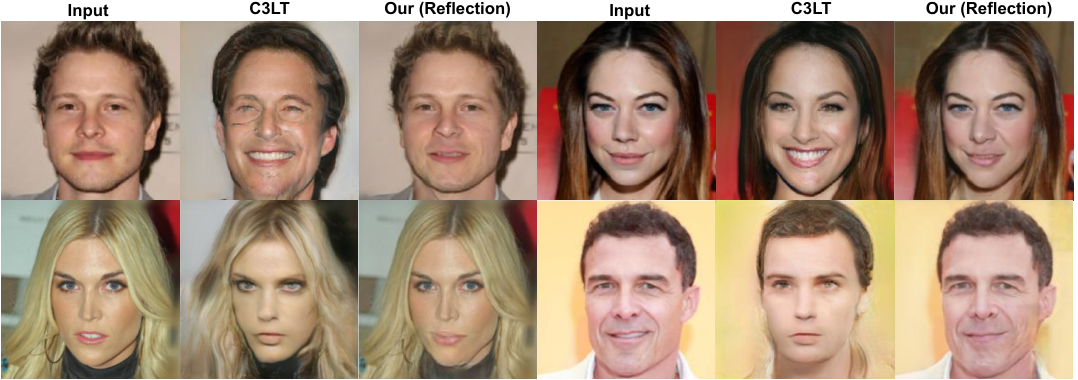}
    \caption{Qualitative comparison between C3LT and Mirror-CFE. The 1st row illustrates closed $\rightarrow$ slightly opened mouth CFEs, while the 2nd row shows the opposite. The image resolutions are 224$\times$224, please zoom in for the details.}
    \label{fig:qualitative_celeba}
\end{figure}

\textbf{Comparison on CelebA-HQ}. First, we utilize ProGAN~\cite{karras2017progressive} as a pre-trained decoder required by C3LT. 
We also omit the comparison to the rest of the methods as they were unable to produce meaningful CFEs for high-resolution images. 
In Fig.~\ref{fig:qualitative_celeba}, we show that C3LT can generate the correct mouth openness in their CFEs but not necessarily preserve the person's identity. 
However, with the control to skip connections, Mirror-CFE can apply subtle, localized changes on the CFEs to realize the intended mouth openness while preserving personal identity. 
Please refer to Sec. \ref{sec:quant_celeba} in the supplementary for quantitative comparison.

\textbf{CFE transition animation.}
We enclose several animated CFE transition videos as multimedia supplementary material to illustrate Mirror-CFE's capability of generating CFEs with pre-determined target class confidence from $k=0$ to $k=1$.

\subsection{Quantitative Evaluations}
\textbf{Metric definition.} 
We evaluate all the methods based on the six common evaluation metrics below. 
\begin{enumerate}
    \item \textit{Validity (Val.):} the classification accuracy towards the target class $\argmax_c {\rho(F(\mathbf{x}_{cf}))} = t$, where $F$ is the trained ResNet classifier.
    \item \textit{Proximity:} the L1 distance between the CFE and source $|\mathbf{x}_{cf} - \mathbf{x}_s|$. 
    \item \textit{Realism:} we employ the Fréchet Inception Distance (FID) \cite{heusel2017gans} to measure realism. We train one autoencoder each for MNIST/F-MNIST to compare all methods and use Inception v3 \cite{szegedy2016rethinking} for B-MNIST and CelebA-HQ.
    \item \textit{Sparsity:} we use Learned Perceptual Image Patch Similarity (LPIPS)~\cite{zhang2018unreasonable} to measure sparsity. 
    LPIPS essentially computes the similarity between the activations of two image patches for a pre-trained network. Here, we employ a pre-trained SqueezeNet \cite{koonce2021squeezenet}. This measure has been shown to match human perception well. A low LPIPS score indicates that image patches between two images are perceptually similar.
    \item \textit{Denoised validity (D.Val)}: to detect the adversarial example generation, we add Gaussian blur to the generated CFE to destroy the imperceptible adversarial changes and re-measure the validity. A higher drop in validity indicates a more likely adversarial generation.
\end{enumerate}

\textbf{Quantitative analysis.} 
The quantitative comparison is shown in Table~\ref{tab:quantitative}. 
As a general observation, the L1, LPIPS, and FID metrics (for proximity, realism, and sparsity respectively) have strong positive correlations, meaning that they rank the reported methods similarly. 
Interestingly, REVISE~\cite{dhurandhar2018explanations} and CEM~\cite{joshi2021decompose} occasionally do not converge (as noted by \% Fail) but achieve the most favorable L1, LIIPS, and FID scores.
Link to the observations from Fig.~\ref{fig:qualitative}, such favorable scores are possible outcomes of adversarial example generation.
On the other hand, validity shows a negative correlation to the above distance metrics, where a high validity in ExpGAN, C3LT, and Mirror-CFE methods is commonly associated with high distance measurements.
Finally, comparing the drop of validity to denoised validity, Mirror CFE is significantly less affected by the Gaussian blur compared to C3LT and ExpGAN.

\textbf{Faithfulness measurement.}
We propose two interrelated faithfulness measurements to describe how closely a generated CFE can achieve the intended CFE feature $\mathbf{z}_k$ and the prediction $\mathbf{p}_k$, \ie, feature reconstruction distance $\mathcal{L}_\text{fea}$ as shown in Eq.~(\ref{eq:fea}) and absolute confidence difference, \ie, $\mathcal{L}_\text{conf\_L1} = |\sigma(\mathbf{z}_k) - \sigma(G(\mathbf{z}_k))|$.
Given that \textit{only} our Mirror-CFE considers the faithfulness, \ie, possesses the ability to calculate an intended classifier feature and target class confidence, only Mirror-CFE in shown in Table~\ref{tab:faithfulness} in the supplementary material.

\subsection{Ablation study}

We show two ablations in Table~\ref{tab:loss_impact} in the supplementary material.
\textit{(1) Different $\alpha$ values.}
In brief, $\alpha = 0$ characterizes the exact preservation of the distance relationships in $\mathcal{Z}$ to $\mathcal{I}$, whereas $\alpha = 0.6$ is the most relaxed. 
Judging from the comparison, $\alpha=0$ over-assumes the relationship, which harms validity, especially for $k = 1$; on the other hand, large $\alpha$ is overindulgent.
Distance metric-wise, the change of $\alpha$ values has a marginal effect.
\textit{(2) With or without triangulation loss.}
$\alpha = 0.2$ is uniformly used in our triangulation loss for all experiments.
Without using the triangulation loss, the measurements are close to using an $\alpha = 0.6$, which is reasonable as having no triangulation loss is similar to an overly passive constraint. \\
We further present an ablation of the skip-connection and CSP in Table~\ref{tab:ablation_ssc} and Fig.~\ref{fig:skip_ablation} (supplementary). The Skip-connection variant closely matches the source image, indicating minimal change. In contrast, removing CSP (No-CSP) causes global alterations. The full SSC module yields localized, semantically meaningful counterfactual edits.

\section{Conclusion}

We present Mirror-CFE, a novel counterfactual explanation method that fundamentally reimagines how CFEs should be generated for deep image classifiers. By operating directly in the classifier's feature space and treating decision boundaries as mirrors, our method ensures faithfulness to the classifier's learned representations and decision-making process. 
Our method uniquely enables visualization of the complete transformation process from source to counterfactual, providing deeper insights into the classifier's behavior. 
While we validated Mirror-CFE on ResNet architectures, future work will explore adaptations to Vision Transformers and other modern architectures (see discussion Sec.~\ref{sec:limitation} in the supplementary material).

{
    \small
    \bibliographystyle{ieeenat_fullname}
    \bibliography{main}
}

\clearpage
\setcounter{page}{1}
\maketitlesupplementary

\section{Mirror CFE for multi-class classification network}
\label{sec:multiclass_mirror_cfe}
\begin{figure}[!h]
    \centering
    \includegraphics[width=0.9\linewidth]{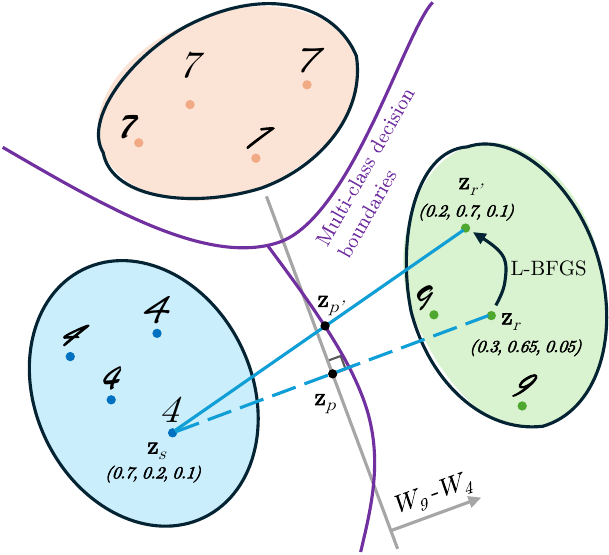}
    \caption{Illustration of the multi-class reflection point calculation that depicts a 3-class toy problem. The introduction of class 7 led to non-linear classification decision boundaries such as shown in the purple color. The source (class 4) and target (class 9) computed binary classification boundary will unlikely to provide a reflection point that flips the class confidence between class 4 and 9. Therefore, we use L-BFGS to estimate a location $\mathbf{z}_{r'}$ which flips the confidence. The confidence are noted as, \eg, (0.2, 0.7, 0.1) for classes 4, 9, and 7 respectively.}
    \label{fig:multiclass}
\end{figure}

\begin{figure*}[!htbp]
    \centering
    \includegraphics[width=1.0\textwidth]{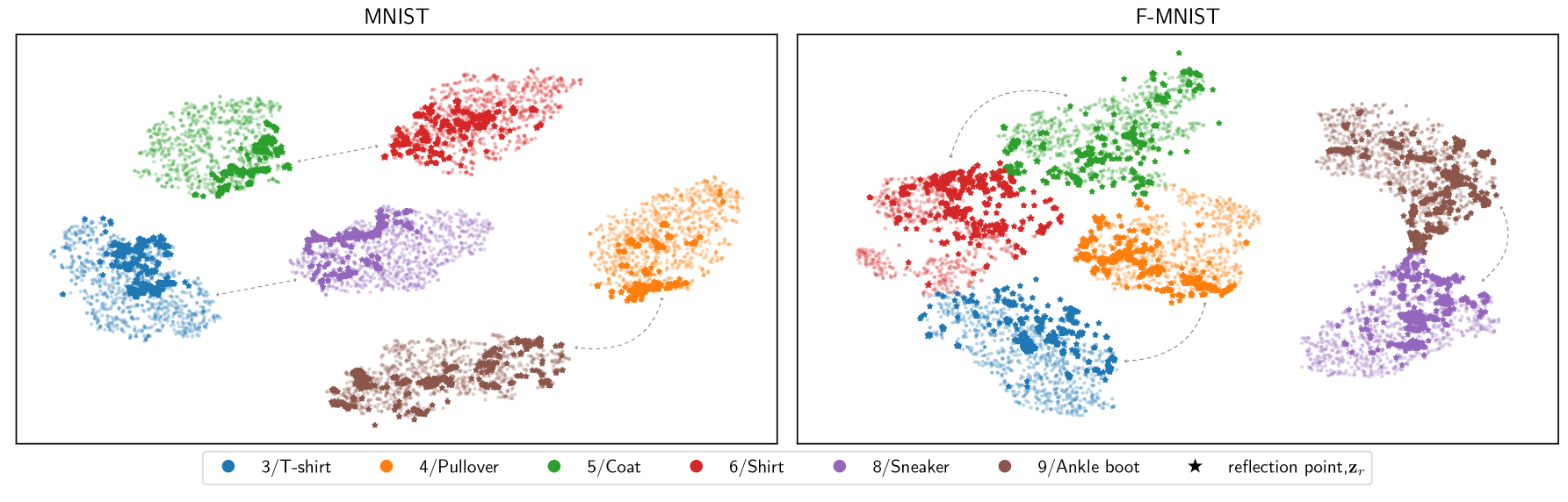}
    \caption{T-SNE visualization of MNIST (left) and F-MNIST (right). Each cluster represents a specific class, with stars marking our computed reflection points ($\mathbf{z}_{r'}$) in the latent feature space. The dashed arrow represents the class pair of source and target class in each dataset. The reflection points are well located with each class cluster, demonstrating their realism within the latent feature space.}
    \label{fig:tsne}
\end{figure*}

\begin{figure}[!htbp]
    \centering
    \includegraphics[width=1.0\columnwidth]{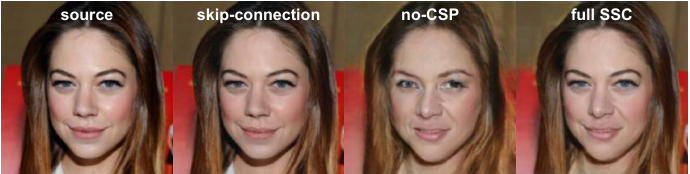}
    \caption{Ablation of the SSC module on CelebA-HQ. }
    \label{fig:skip_ablation}
\end{figure}

$\mathbf{q}_{(cf, t)} > \mathbf{q}_{(cf, s)}$ for $k = 0.5 + \epsilon$ is only true for a binary classification setting.
It may not hold in a multi-class classification setting due to interference from the other classes in the softmax probability calculation.
While the exact $\mathbf{z}_p$ and $\mathbf{z}_r$ positions are not directly computable in the multi-class classification setting, we use Eq.~(\ref{eq:position}) obtained $\mathbf{z}_r$ to initialize the L-BFGS~\cite{byrd2016stochastic} algorithm with the optimization goal that a new $\mathbf{z}_{r'}$ location will satisfy $\min_{\mathbf{z}_{r}} \|\mathbf{l}_{r} - \mathbf{l}_{r'}\|,$.
This process is depicted in Fig.~\ref{fig:multiclass}.
The reflection point's multi-class logits are calculated as $\mathbf{l}_{r} = \mathbf{W}^\top \mathbf{z}_{r} + \mathbf{b}$ and $\mathbf{l}_{r'}$ a hypothetical reflection point's logit that flips the confidence between the source and target classes: $\mathbf{l}_{(r',s)} = \mathbf{l}_{(s, t)}$  and $\mathbf{l}_{(r',t)} = \mathbf{l}_{(s, s)}$, the rest class logits remain unchanged.
For the KFE points on the line from $\mathbf{z}_s$ to $\mathbf{z}_{r'}$ the computation of their location is the fraction of the travel indicated by $k$.
For example, we calculate the new projection point $\mathbf{z}_{p'} = \frac{\mathbf{z}_s + \mathbf{z}_{r'}}{2}$.
Note that $\mathbf{z}_{p'}$ can achieve $\mathbf{p}_{({p'}, s)} \approx \mathbf{p}_{({p'}, t)}$, but $\mathbf{p}_{({p'}, s)}, \mathbf{p}_{({p'}, t)} \neq 0.5$ due to the presence of other classes.
Finally, we found the L-BFGS can find reasonably accurate logits that reflect the intended probabilities.
In Fig. \ref{fig:tsne}, we show our generated multi-class classification reflection points in $\mathcal{Z}$ by using T-SNE visualization. 
It can be seen that the generated reflection points are correctly located in the T-SNE computed clusters. \\
Our Mirror-CFE supports multi-class training, but we followed the binary settings for fair comparison with baselines. Multi-class results in Table~\ref{tab:multi-class} are similar to the pairwise counterparts. CelebA-HQ has binary attributes, hence there is no difference using multi-class.

\begin{table}[!hbp]
\centering
\resizebox{1.0\columnwidth}{!}{
\begin{tabular}{c|c|c|c|c|c|c}
\toprule
 & & L1$\downarrow$ & LPIPS$\downarrow$ & FID $\downarrow$ & D.Val.$\uparrow$ & Val.$\uparrow$\\
\midrule
 \multirow{2}{*}{\rotatebox[origin=c]{90}{{\tiny MNIST}}} & 
Pairwise & \bf 0.26 & \bf 0.33 & \bf 3.20 & \bf 0.99 & \bf 1.0\\
 & Multiclass & 0.27 & 0.34 & 6.94 & \bf 0.99 & \bf 1.0\\
\midrule
\multirow{2}{*}{\rotatebox[origin=c]{90}{{\tiny FMNIST}}} & 
Pairwise  & 0.25  & 0.21 & \bf 4.40 & 0.96 & \bf 0.99\\
 & Multiclass & \bf 0.20 & \bf 0.19 & 5.83 & \bf 0.97 & 0.98\\
\midrule
\end{tabular}
}
\caption{Comparison between Multi-class and pairwise settings of our Mirorr-CFE on MNIST and Fashion-MNIST.}
\label{tab:multi-class}
\end{table}

\section{Computing KFE encoder feature from latent space encoding}
\label{sec:kfe_enc_from_latent}
Given a KFE latent code $\mathbf{z}_k$ can be computed by using Eq.~(\ref{eq:position}) for any $k$ from $\mathbf{z}_s$, we show how $\mathbf{f}_k^l$ is computed from $\mathbf{z}_k$:
\begin{equation}
\mathbf{z}_s = \text{GAP}(\mathbf{f}^i_k) = \sum_{h,w\in H_l, W_l} \frac{\mathbf{f}^l_s}{H_l\times W_l}.
\end{equation}
Assuming $\mathbf{z}_k = \sum_{h,w\in H_l, W_l} \frac{\mathbf{f}^l_k}{H_l\times W_l}$ was also computed from the GAP, and using Eq.~\ref{eq:position} to calculate $\mathbf{z}_{k}$:
\begin{align}
\mathbf{z}_k  = & \mathbf{z}_s - 2k (\mathbf{W}^\top_m \mathbf{z}_s + \mathbf{b}_m) \hat{\mathbf{W}}_m.
\end{align}
Let $\mathbf{z}_{\Delta} = - 2k (\mathbf{W}^\top_m \mathbf{z}_s + \mathbf{b}_m) \hat{\mathbf{W}}_m$, we show the equality:
\begin{align}
\sum_{h,w\in H_l, W_l} \frac{\mathbf{f}^l_k}{H_l\times W_l}  = & \sum_{h,w\in H_l, W_l} \frac{\mathbf{f}^l_s}{H_l\times W_l} + \mathbf{z}_{\Delta}, \nonumber\\
= & \sum_{h,w\in H_l, W_l} \frac{\mathbf{f}^l_s + \mathbf{z}_{\Delta}}{H_l\times W_l} 
\end{align}
hence, $\mathbf{f}^l_k  = \mathbf{f}^l_s + \mathbf{z}_{\Delta}$ is a trivial solution that assumes all elements in $\mathbf{f}^l_k$ take the same rate of travel from $\mathbf{f}^l_s$.

For the multi-class classification scenario, the calculation is slightly different.
We first follow Sec.~\ref{sec:multiclass_mirror_cfe} to find the updated position of the reflection point $\mathbf{z}_{r'}$.
Then we compute $\mathbf{z}_\Delta$ differently  as $\mathbf{z}_\Delta = k(\mathbf{z}_r - \mathbf{z}_s)$ and finally $\mathbf{f}^l_k  = \mathbf{f}^l_s + k(\mathbf{z}_r - \mathbf{z}_s)$.

\section{Quantitative Analysis of CelebA-HQ}
\label{sec:quant_celeba}

The quantitative comparison between C3LT and Mirror-CFE is in Table~\ref{tab:quant_celeba} for CelebA.
In general, both our 1st CFE and mirror CFE $(k-1)$ show significant improvement in L1, LPIPS, FID, and denoised validity in comparison to C3LT.
For validity, Mirror-CFE obtained 0.94, and C3LT achieved 1. We note that the Mirror-CFE model was trained with 30 epochs, while 200 epochs were used for C3LT.

\begin{table}[!htbp]
\centering
\resizebox{0.85\columnwidth}{!}{
\begin{tabular}{c|c|c|c|c|c}
\toprule
Method & L1$\downarrow$ & LPIPS$\downarrow$ & FID $\downarrow$ & D.Val.$\uparrow$ & Val.$\uparrow$\\
\midrule
C3LT & 0.21 & 46.41 & 35.28 & 0.81 & \bf 1.0\\
Our (1st CFE) & \bf 0.08 & \bf 0.051 & \bf 8.78 & 0.87 & 0.94\\
Our ($k=1$) & \bf 0.08 & 0.053	& 8.97 & \bf 0.89 & 0.94\\
\bottomrule
\end{tabular}
}
\caption{Quantitative analysis of CelebA-HQ.}
\label{tab:quant_celeba}
\end{table}

\section{Description of the animation files}

\subsection{Folder structure}
We provide a collection of CFE transition animation videos, organized into dataset folders containing subfolders for specific class pairs\footnote{\url{https://github.com/AIML-MED/Mirror-CFE/tree/main/suppelementary/animations}}. 
In the MNIST dataset, sub-folders 38, 49, and 56 represent the class pairs 3 vs. 8, 4 vs. 9, and 5 vs. 6, respectively. For the F-MNIST dataset, sub-folders 02, 46, and 79 correspond to the class pairs T-shirt vs. Pullover, Coat vs. Shirt, and Sneakers vs. Boots. For the B-MNIST dataset, subfolder 24 represents the class pair Erythroblast vs. Lymphocyte. 
For CelebA-HQ, subfolder 01 represents the attribute pair mouth closed (0) vs. slightly mouth open (1).
Each subfolder includes an MP4 file demonstrating the CFE transition for the examples shown in Fig.~\ref{fig:qualitative} or Fig.~\ref{fig:qualitative_celeba}. 

\subsection{File naming}
The file naming convention for these animations is \textit{fig3\_rowN\_A\_(GT\_B)\_to\_C\_\{correct/error\}.mp4}, and the naming convention is:
\begin{itemize}
    \item rowN is the row number a sample was placed in Fig.~\ref{fig:qualitative}, or any combination of top/bottom and left/right for samples in Fig.~\ref{fig:qualitative_celeba}.
    \item A is the model-predicted label, we use it as the source class for the CFE generation. A is not necessarily equal to B because a model may give error prediction.
    \item B is the ground truth (GT) label for reference.
    \item C is the target class label for the CFE generation.
    \item `correct' means this case was predicted correctly, \ie, A = B, or `error' otherwise.
\end{itemize}
We also show additional examples in a folder called ``additional samples''.
The naming convention is similar but changes the `fig3\_rowN' prefix to `noN' indicating a unique file identifier (otherwise the file names could be duplicated).

\begin{table}[!tbp]
\centering
\resizebox{0.55\columnwidth}{!}{
\begin{tabular}{c|c|c|c}
\toprule
 & MNIST & F-MNIST &  B-MNIST\\
\midrule
$\mathcal{L}_\text{fea}$ & 0.012 & 0.013 & 0.085\\
$\mathcal{L}_\text{conf\_L1}$ & 0.014 & 0.024 & 0.015\\
\bottomrule
\end{tabular}
}
\caption{The feature reconstruction (L1 distance) and KLD measurements of faithfulness to the anticipated CFE feature and target confidence for Mirror-CFE. In brief, Mirror-CFE achieves on average $< 3\%$ confidence difference to intended target confidence.}
\label{tab:faithfulness}
\end{table}

\begin{table}[!tbp]
\centering
\resizebox{0.85\columnwidth}{!}{
\begin{tabular}{c|c|c|c|c|c}
\toprule
& L1$\downarrow$ & LPIPS$\downarrow$ & FID $\downarrow$ & D.Val.$\uparrow$ & Val.$\uparrow$\\
\midrule

Skip-Connection & \bf 0.07 & \bf 0.02 & \bf 4.20 & 0.03 & 0.03 \\
Ours (no-CSP)    & 0.14 & 0.19 & 40.12 & 0.78 & 0.78 \\
Ours (full-SSC) & 0.08 & 0.05	& 8.97 & \bf 0.89 & \bf 0.94\\
\midrule
\end{tabular}
}
\caption{Ablation study of SSC and CSP on CelebA-HQ. The `Skip-connection' setting denotes the `no-SPE' variant, where $\mathbf{u}^i_k = \mathbf{f}^i_s$ as defined in Eq.~\ref{eq:mixture}.}
\label{tab:ablation_ssc}
\end{table}

\begin{table}
\centering
\resizebox{0.85\columnwidth}{!}{
\begin{tabular}{c|c|c|c|c|c|c}
\toprule
$\mathcal{L}_\text{tri}$ & $k$ & L1$\downarrow$ & LPIPS$\downarrow$ & FID $\downarrow$ & D.Val.$\uparrow$ & Val.$\uparrow$\\
\midrule

$\alpha = 0$ & $\neq 1$ & 0.126 & 0.104 & 2.302 & 0.989 & 0.991\\
$\alpha = 0.2$  & $\neq 1$ & 0.127 & 0.109 & 2.801 & 0.990 & 0.989\\
$\alpha = 0.4$ & $\neq 1$ & 0.128 & 0.107 & 3.001 & \bf 0.992 & 0.990\\
$\alpha = 0.6$ & $\neq 1$ & 0.116 & 0.102 & \bf 2.141 & 0.989 & 0.987 \\
\xmark & $\neq 1$ & \bf 0.115 & \bf 0.099 & 2.469 & 0.996 & \bf 0.994\\
\midrule
$\alpha = 0$ & 1 & \bf 0.185 & \bf 0.190 & \bf 3.364 & 0.955 & 0.905\\
$\alpha = 0.2$ & 1 & 0.251 & 0.217 & 4.402 & \bf 0.960 & 0.949\\
$\alpha = 0.4$ & 1 & 0.260 & 0.224 & 4.510 & 0.915 & 0.965\\
$\alpha = 0.6$ & 1 & 0.256 & 0.221 & 4.262 & 0.906 & 0.956\\
\xmark & 1 & 0.248 & 0.215 & 4.227 & 0.921 & \bf 0.971\\

\bottomrule
\end{tabular}
}
\caption{Ablation of $\mathcal{L}_\text{tri}$ loss on F-MNIST dataset, separately computed for the CFE computed from the reflection point ($k=1$) and the rests ($k\neq1$) (1st CFE to any CFEs with $k<1$).}
\label{tab:loss_impact}
\end{table}

\begin{table*}[!bp]
    \centering
    \begin{tabularx}{\linewidth}{lX}
    \toprule
    Symbol & Meaning/Definition \\
    \midrule
    $\mathcal{X}, \mathcal{C}, \mathcal{Z}, \mathcal{I}$ & These represent the image set, label set, latent feature space, and image (pixel) space, respectively.\\ \hline
    $|\mathcal{C}|$ (scalar) & The size of the label set. \\ \hline
    $s, t$ (scalars)  & Source and target classes.\\ \hline
    $\mathbf{x} \in \mathbb{R}^{C \times H \times W}, y$ \text{ (scalar)}& An image $\mathbf{x}$ associated with the class label $y$. \\ \hline
    $F(\mathbf{x})$ & The classifier model.\\ \hline
    $\mathbf{z} \in \mathbb{R}^N$ & Latent feature representation $\mathbf{z}$ with the size $N$ (\eg, 2048) computed as $\mathbf{z} = F(\mathbf{x})$.\\ \hline
    $C$, $H$, $W$ & The number of input channels and spatial dimensions of the input image.\\ \hline
    $\mathbf{W} \in \mathbb{R}^{N \times |\mathcal{C}|}, \mathbf{b} \in \mathbb{R}^{|\mathcal{C}|}$ & Weight and bias of the classifier $F$'s classification layer respectively.\\ \hline
    $\mathbf{p} \in \mathbb{R}^{|\mathcal{C}|}, \sigma(\mathbf{z})$ & Probability distribution $\mathbf{p}$ over classes, computed as $\mathbf{p} = \sigma(\mathbf{z}) = \text{softmax}(\mathbf{W}^\top\mathbf{z} + \mathbf{b})$ function applied to the latent feature representation $\mathbf{z}$.\\ \hline
    $\mathbf{l} \in \mathbb{R}^{|\mathcal{C}|}$ & Logit before applying softmax, \ie, $\mathbf{l} = \mathbf{W}^\top\mathbf{z} + \mathbf{b}$.\\ \hline
    $\mathbf{W}_s, \mathbf{W}_t \in \mathbb{R}^{N}; \mathbf{b}_s, \mathbf{b}_t$ (scalars) & $\mathbf{W}_s$  and $\mathbf{W}_t$ are weight vectors which are the ``slices'' of $\mathbf{W}$ for the classes $s$ or $t$. Similarly, $\mathbf{b}_s$ and $\mathbf{b}_t$ are the elements in the bias vector $\mathbf{b}$ for classes $s$ and $t$.\\ \hline
    $\mathbf{W}_m \in \mathbb{R}^{N}, \mathbf{b}_m$ (scalar) & Pairwise decision boundary between the source and target classes, \ie, the mirror. Calculated as the difference of class weights ($\mathbf{W}_t - \mathbf{W}_s$) and bias ($\mathbf{b}_t - \mathbf{b}_s$).\\ \hline
    $G(\mathbf{z})$ & A mapping function $G$ to project latent feature $\mathbf{z}$ into image space.\\ \hline
    $k$ (scalar) & Step factor where $k \in [0,1]$.\\ \hline
    $\mathbf{f}^{i} \in \mathbb{R}^{C_{i}\times H_{i} \times W_{i}}$ & A spatial feature $\mathbf{f}^{i}$ is obtained from $i$-th layer of the feature encoder $F$. $C_i, H_i$ and $W_i$ denote the channel, height, and width of $\mathbf{f}^i$. \\ \hline
    $l$ (scalar) & The highest layer in $F$ which produces $\mathbf{z}$: $\mathbf{z} = \text{GAP}(\mathbf{f}^l)$, hence $N = C_l$. \\ \hline
    $\mathbf{x}_{ss}, \mathbf{z}_{ss}$ & A random source class image $\mathbf{x}_{ss}$ corresponds to latent feature $\mathbf{z}_{ss}$.\\ \hline
    $\mathbf{x}_{r}, \mathbf{z}_{r} $ & A generated mirror point image $\mathbf{x}_{r}$ from mirror latent feature $\mathbf{z}_{r}$ at step factor $k=1.0$.\\ \hline
    $\mathbf{x}_{cf}, \mathbf{z}_{cf} $ & A generated counterfactual image $\mathbf{x}_{cf}$ from  latent feature $\mathbf{z}_{cf}$ at step factor $k = 0.5 + \epsilon$ where $0<\epsilon <=0.5$.\\ \hline
    $\mathbf{x}_{sf}, \mathbf{z}_{sf} $ & A generated semi-factual image $\mathbf{x}_{sf}$ from latent feature $\mathbf{z}_{sf}$ at step factor $k = 0.5 - \epsilon$.\\ \hline
    $\mathbf{x}_{k}, \mathbf{z}_{k} $ & A generated KFE image $\mathbf{x}_{k}$ from latent feature $\mathbf{z}_{k}$ at step factor $k \in [0,1]$.\\ \hline
    $\mathcal{L}_{cls}$	& Classification loss ensuring the validity of counterfactual explanations.\\ \hline
    $\mathcal{L}_{adv}$	& Adversarial loss ensuring realism in generated counterfactuals.\\ \hline
    $\mathcal{L}_{rec}$	& Reconstruction loss ensuring regenerated images resemble the original inputs.\\ \hline
    $\mathcal{L}_{fea}$	& Feature reconstruction loss for consistency in latent space.\\ \hline
    $\mathcal{L}_{tri}$	& Triangulation loss maintaining proximity and realism during counterfactual generation.\\ \hline
    $\alpha$ (scalar) & Scaling factor in triangulation loss $\mathcal{L}_{tri}$ where $\alpha \in [0, 1]$.\\ \hline
    $B_i, D_i$ & The bottleneck and decoder modules used in the $i$-th layer's Spatial pattern editor (SPE) module. \\ \hline
    $\mathbf{u}^i_k \in \mathbb{R}^{C_{i}\times H_{i} \times W_{i}}$ & Output from the SPE module in $i$-th layer at step $k$.\\ \hline
    $\mathbf{U}_k \in \mathbb{R}^{|\mathcal{C}|\times H_l \times W_l}$ & Unnormalized CAMs at step $k$ computed as $\mathbf{U}_k = \mathbf{W}^\top \mathbf{f}^{l}_{k}$. \\ \hline
    $\mathbf{N}_k \in \mathbb{R}^{|\mathcal{C}|\times H_l \times W_l}$ & Normalized CAMs at step $k$ where each channel is individually normalized between [0, 1]. \\ \hline
    $\mathbf{M}^i_k \in \mathbb{R}^{H_{i} \times W_{i}}$ & Spatial prior mask $i$-th layer at step $k$. $\mathbf{M}^i_k$ is uniformly applied to all channels.\\
    \bottomrule
    \end{tabularx}
    \caption{Symbol lookup table.}
    \label{tab:symbol_lookup}
\end{table*}

\subsection{Video content}
Each video is structured into three sections per time frame. 

The \textbf{top section} displays the Source image ($\mathbf{x}_s$) alongside the KFE image ($\mathbf{x}_k$). The displayed label of source image is `Source($\argmax{\mathbf{p}_s}$)' and the KFE image label is one of `SFE/CFE/Reflection($\argmax{\mathbf{p}_k}$)'. 

The \textbf{middle section} visualizes the transition of the image in the latent feature space, moving from the source latent point ($\mathbf{z}_s$) to the reflection point ($\mathbf{z}_r$). 
The first significant transition point to obtain CFE image $\mathbf{x}_{cf}$ is highlighted in this section as the ``1st CFE point". 
The 1st CFE means the transition first finds an image where the model predicts the image as a target sample.
Additionally, the intended and predicted confidence levels for both the source and target classes are displayed as text `(S:, T:)' in orange and blue, respectively. 

The \textbf{bottom section} presents a confidence plot for the target class. 
This plot also includes the difference between the source image and the KFE image (\ie, $\frac{1}{C\times H\times W} \sum |\mathbf{x_k} - \mathbf{x_s}|$) at each step $k$ (green line). 
This plot provides a comprehensive visualization of the transformation process across different datasets and class pairs, showing the $k$ where the most image content change happens and how it affects the classifier's prediction (in confidence).\\

\subsection{Additional samples}

We provide additional examples to showcase the efficacy of our method. Across the MNIST dataset, all examples exhibit a smooth and meaningful transition from the source label to the target label, demonstrating the effectiveness of Mirror CFE. 
In F-MNIST, the transitions vary in magnitude depending on the class pair. 
For instance, the changes between shirt and coat\footnote{fmnist/46/additional samples} are minimal, while the transformations between sneaker and ankle boot\footnote{fmnist/79/additional samples} are considerably more pronounced. 
The transitions for the T-shirt and pullover\footnote{fmnist/02/additional samples} class pair consistently occur in the sleeve region, aligning with human expectations. 
In B-MNIST\footnote{bmnist/24/additional samples}, we observe that an erythroblast nucleus is generally larger than a lymphocyte. 
On the other hand, lymphocyte has a smaller amount of cytoplasm than erythroblast.
This distinction is also observed in the transition process for this class pair in our examples, further validating Mirror CFE's capability to visualize meaningful class-wise differences.

\subsection{`Error' cases}
For the error cases, the source image was not predicted as its GT source label ($\argmax{p_s} \neq s$). 
This is an instinct learning problem of the classifier, not because of Mirror CFE.
In such instances, Mirror CFE presents a unique feature to analyze how to rectify the error cases, which are unavailable from the compared methods.

When the source image is misclassified we take the predicted class as the source label and set $t$ as the ground truth class. 
For example, we present an error case\footnote{mnist/49/additional samples/no31\_9(GT\_4)\_to\_4\_error.mp4} in the MNIST dataset involving the class pair 4 and 9.%
Here, a source image of the digit 4 is misclassified as digit 9 and we construct the CFE from 9 to 4, highlighting the modifications necessary for it to be classified as 4. 
Specifically, we observe the top portion of the digit 4 bends upwards when $k$ is nearly 0.8, which dramatically increases its confidence as being correctly recognized as a 4 by the classifier.

Another example\footnote{fmnist/46/additional samples/no5\_Shirt(GT\_Coat)\_to\_Coat\_error.mp4} can be observed in the F-MNIST dataset, involving the class pair `Shirt' and `Coat'. 
In this case, a coat is misclassified as a shirt. 
Notably, many shirts in F-MNIST feature checkered patterns, which are also present in this specific misclassified sample. Hence, the classifier was making a reasonable assumption that it is a shirt. 
Our method `cleaned' the checkered patterns in the image, showing users what it takes to make the image correctly classified as a coat, giving users a more intuitive understanding of the class difference, specifically illustrated in this sample.

\section{Limitations}
\label{sec:limitation}
Mirror-CFE requires the existence of a Global Average Pooling (GAP) layer in the classifier's architecture to facilitate the functionality of the CAM-guided spatial prior (CSP) module.  
We choose Class Activation Maps (CAM) because CAM is a feed-forward computation that can be computed in the same feed-forward iteration--much faster than gradient-based and sampling-based attention methods.
We acknowledge that GAP-enabled models are rare in Transformers, nor do they provide meaningful CAMs off-the-shelf; hence, the usage of Mirror-CFE would require heuristic implementation to estimate the network attention, such as modeling the Transformer attention via \cite{abnar2020quantifying, chefer2021transformer}.

\end{document}